\documentclass[runningheads]{llncs}
\usepackage[T1]{fontenc}
\usepackage{graphicx,verbatim}
\usepackage{amssymb}
\usepackage{graphicx,verbatim}
\usepackage{listings}
\usepackage{bm}
\usepackage{color,colortbl}
\usepackage{booktabs} 
\usepackage{multirow} 

\usepackage[colorlinks=true, allcolors=blue]{hyperref}
\definecolor{Gray}{gray}{0.9}
\usepackage{float}
\usepackage{orcidlink}
\usepackage{arydshln}

\usepackage[table]{xcolor}
\definecolor{seagreen}{RGB}{76, 175, 80}
\definecolor{seablue}{RGB}{33, 150, 243}
\definecolor{seaorange}{RGB}{255, 152, 0}

\begin{document}
\title{No Image, No Problem: End-to-End Multi-Task Cardiac Analysis from Undersampled k-Space}
\titlerunning{k-MTR}
%
\author{Yundi Zhang\inst{1}\orcidlink{0009-0008-7725-6369} \and
Sevgi Gokce Kafali\inst{1}\orcidlink{0000-0001-5941-5399} \and
Niklas Bubeck\inst{1}\orcidlink{2222--3333-4444-5555} \and
Daniel Rueckert\inst{1,2,3}\orcidlink{0000-0002-5683-5889} \and
Jiazhen Pan\inst{1}\orcidlink{0000-0002-6305-8117}}

\authorrunning{Y. Zhang, et al.}
%
\institute{Chair for AI in Healthcare and Medicine, Technical University of Munich (TUM) and TUM University Hospital, Munich, Germany \and
Department of Computing, Imperial College London, United Kingdom \and 
Munich Center for Machine Learning, Technical University of Munich, Germany
\\
\email{\{yundi.zhang, jiazhen.pan\}@tum.de}}


  
\maketitle              
\setcounter{footnote}{0}
\begin{abstract}

Conventional clinical CMR pipelines rely on a sequential "reconstruct-then-analyze" paradigm, forcing an ill-posed intermediate step that introduces avoidable artifacts and information bottlenecks. This creates a fundamental mathematical paradox: it attempts to recover high-dimensional pixel arrays (i.e., images) from undersampled k-space, rather than directly extracting the low-dimensional physiological labels actually required for diagnosis. To unlock the direct diagnostic potential of k-space, we propose k-MTR (k-space Multi-Task Representation), a k-space representation learning framework that aligns undersampled k-space data and fully-sampled images into a shared semantic manifold. Leveraging a large-scale controlled simulation of 42,000 subjects, k-MTR forces the k-space encoder to restore anatomical information lost to undersampling directly within the latent space, bypassing the explicit inverse problem for downstream analysis. We demonstrate that this latent alignment enables the dense latent space embedded with high-level physiological semantics directly from undersampled frequencies. Across continuous phenotype regression, disease classification, and anatomical segmentation, k-MTR achieves highly competitive performance against state-of-the-art image-domain baselines. By showcasing that precise spatial geometries and multi-task features can be successfully recovered directly from the k-space representations, k-MTR provides a robust architectural blueprint for task-aware cardiac MRI workflows.
\footnote{Code and models will be made publicly available upon publication.}


\end{abstract}
\section{Introduction}
\label{intro}

In Cardiac Magnetic Resonance (CMR) imaging, the sensor data (k-space) and the spatial information (i.e., images) are dual representations of the same anatomical and physiological reality, mathematically linked by the Fourier Transform. Due to hardware limits, high costs, and requirements for breath-hold by patients, standard cardiac MRI protocols typically undersample k-space to accelerate acquisitions~\cite{bluemke1997hold,wang2001motion,plein2021we}. Thus, the conventional research community has viewed the relationship between these domains primarily through the lens of image reconstruction. The standard objective has been to recover high-fidelity images from undersampled k-space, which inevitably introduces avoidable information bottlenecks and reconstruction artifacts~\cite{seitzer2018adversarial,dohmen2025similarity}.

While deep learning has driven significant progress in image reconstruction~\cite{schlemper2017deep,pan2023global,huang2025subspace}, explicit reconstruction serves merely as an intermediate step. The ultimate clinical goal of CMR is to extract physiological features, clinical phenotypes, and disease labels~\cite{chen2020deep,martin2020image,liu2024review}. This creates a paradox in the standard clinical pipeline. Reconstructing a fully sampled image from undersampled k-space is fundamentally an ill-posed problem because it attempts to recover high-dimensional variables (image) from low-dimensional inputs (k-space). However, the downstream clinical classification and phenotyping tasks are inherently dimensionality-reduction problems. Extracting a low-dimensional disease label from undersampled k-space is mathematically much more well-posed than reconstructing the entire image array.

This realization prompts a critical question: \textbf{Can cardiac analysis operate directly within the undersampled k-space?} Recent studies have begun investigating end-to-end approaches that predict single clinical labels or segmentations directly from undersampled k-space~\cite{schlemper2018linet,li2024classification,zhang2024direct}. While promising, these efforts remain largely restricted to isolated, single-task applications. Concurrently, cardiac multi-task foundation models~\cite{zhang2024whole,jacob2025towards,zhang2025vita} have demonstrated the power of dense latent spaces in the image domain. Yet, the field lacks a unified framework that successfully bridges k-space, spatial images, and clinical labels through a shared dense manifold. 

To address this gap, we propose \textbf{k-MTR} (k-space Multi-Task Representation), an end-to-end framework that constructs a comprehensive, information-rich manifold by aligning the undersampled k-space and image domains. We hypothesize that undersampled k-space retains sufficient information to directly inform clinical endpoints. By bypassing the reconstruction step, this unified manifold enables the diverse cardiac analysis tasks from undersampled measurements. Our key contributions are:

\begin{itemize}
    \item \textbf{The first k-space representation learning framework beyond reconstruction.}
    We introduce k-MTR, the first representation learning approach to align undersampled k-space and spatial images into a shared semantic manifold, entirely bypassing the image reconstruction.

    \item \textbf{Information-dense semantic manifold.}
    We demonstrate that this domain alignment creates an information-dense latent space, implicitly compensating for anatomical structures degraded by undersampling while preserving critical diagnostic semantics.

    \item \textbf{Direct multi-task cardiac analysis from undersampled k-space.}
    k-MTR establishes a unified new paradigm for frequency-domain analysis, achieving highly competitive performance across continuous phenotype regression, disease classification, and fine-grained anatomical segmentation.
\end{itemize}

\begin{figure}
\center
\includegraphics[width=\linewidth]{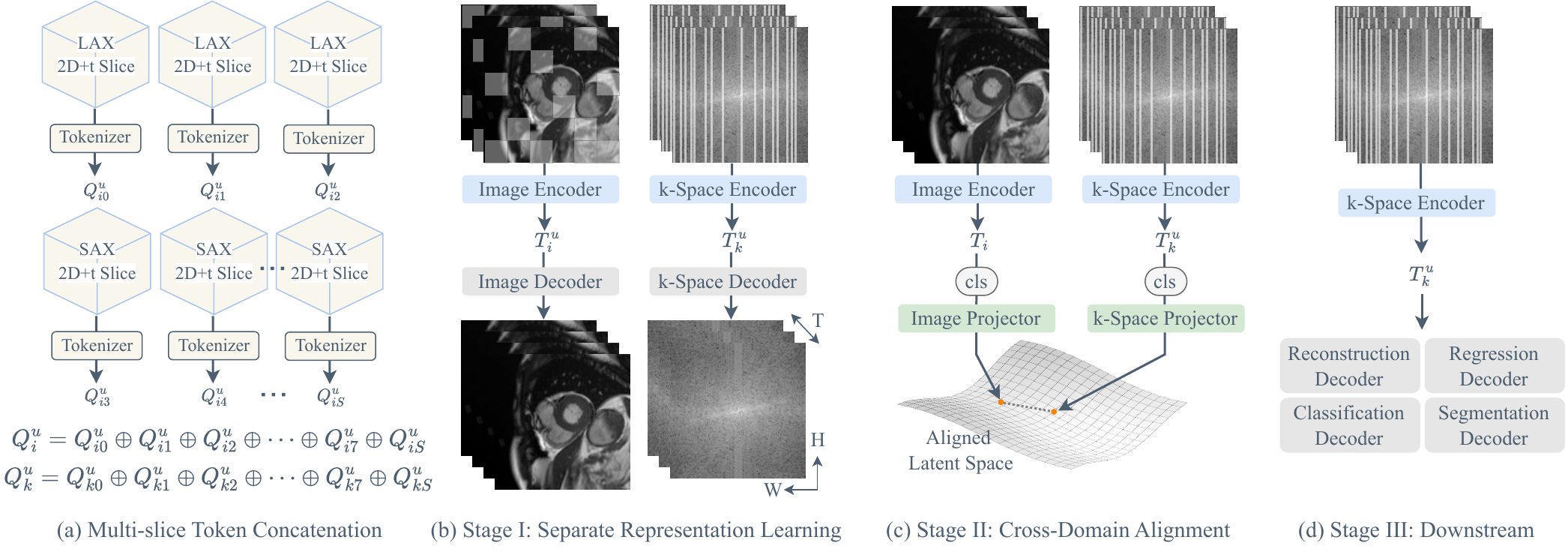}
\caption{\textbf{Overview of the k-MTR framework.} \textbf{(a)} $S$ multi-view $2D+t$ slices are tokenized, concatenating image ($Q_i^u$) and k-space ($Q_k^u$) tokens across slices for encoder input. \textbf{(b--d)} Training pipeline (single slice shown). \textbf{(b)} Unsupervised masked reconstruction of undersampled k-space and image slices. \textbf{(c)} Contrastive alignment between undersampled k-space ($T^u_k$) and fully-sampled image ($T_i$) embeddings. \textbf{(d)} Fine-tuning the pretrained k-space encoder via lightweight task-specific decoders.}
\label{fig:archi}
\end{figure}


\section{k-MTR Methodology}

The three-stage pipeline of k-MTR for effective k-space feature extraction is shown in Fig.~\ref{fig:archi}.
Let fully-sampled complex-valued cardiac k-space measurements be $X_k \in \mathbb{C}^{S\times T\times H\times W}$, where $S, T, H$, and $W$ represent the number of slices, time frames, height, and width, respectively. The slices, comprising both $2D+t$ long-axis (LAX) and short-axis (SAX) views, are concatenated along the slice dimension to yield the corresponding image stack $X_i \in \mathbb{C}^{S\times T\times H\times W}$. To simulate standard clinical acquisition constraints, we apply a Cartesian acceleration mask $M \in \mathbb{Z}^{S\times T\times W}$ along the phase-encoding ($W$) direction. The mask is broadcast along the $H$ dimension to obtain $\tilde{M}$, yielding the undersampled k-space data via element-wise multiplication: $X^u_k = \tilde{M}\odot X_k$. The superscripts 
$i$, $k$, and $u$ denote the image domain, k-space domain, and undersampled data, respectively.

\textbf{Stage I: Domain-Specific Representation Learning.}
In the first stage, we independently learn robust domain-specific features using the masked autoencoder (MAE) paradigm~\cite{he2022masked}. For the image domain, multi-view $2D+t$ images are randomly masked at the patch level to obtain $X^u_i$. For the frequency domain, the k-space data relies on the predefined clinical acceleration mask $\tilde{M}$ to yield $X^u_k$. Each domain employs its own tokenizer $\mathcal{T}$, encoder $\mathcal{E}$, decoder $\mathcal{D}$, and mask tokens $T^m$. With $\oplus$ denoting concatenation, the reconstruction objectives are formulated as:
\begin{equation}
\hat{X}_i = \mathcal{D}_i(\mathcal{E}_i(Q^u_i) \oplus T^m_i), \quad
\hat{X}_k = \mathcal{D}_k(\mathcal{E}_k(Q^u_k) \oplus T^m_k),
\end{equation}
where $Q_i^u = \mathcal{T}_i(X_i^u)$ and $Q_k^u = \mathcal{T}_k(X_k^u)$ denote the tokenized inputs. Real and imaginary components are processed as two input channels, and multi-slice tokens are concatenated along the sequence-length dimension. This ensures that both encoders establish domain-specific semantic capacities, preparing for subsequent cross-modal alignment.

\textbf{Stage II: Cross-Modal Alignment and Latent Restoration.} We establish a shared latent space to align image and k-space representations. A critical design choice is the asymmetry of the inputs: image representations $T_i=\mathcal{E}_i(\mathcal{T}_i(X_i))$ are extracted from \textbf{fully-sampled} data to preserve complete semantic geometry, while k-space representations $T_k^u=\mathcal{E}_k(\mathcal{T}_k(X^u_k))$ are derived solely from \textbf{undersampled} data. It explicitly forces the k-space encoder $\mathcal{E}_k$ to recover and embed the anatomical signatures omitted by the undersampling mask directly into its latent vector, implicitly bypassing the inverse problem in the latent space. We project class tokens via domain-specific projectors $\mathcal{P}_i$ and $\mathcal{P}_k$ to obtain $\hat{Z}_i = \mathcal{P}_i(T_i)$ and $\hat{Z}_k = \mathcal{P}_k(T_k^u)$. With temperature $\tau$ and loss weight $\lambda$, a symmetric contrastive loss $\mathcal{L}$ is applied over a batch of subjects $\mathcal{N}$ to minimize the distance of multi-modal embeddings of the same subject:
\begin{equation}
    \ell_{i, k}=-\sum_{m \in \mathcal{N}} \log \frac{\exp \left(\cos \left(z_{m_i}, z_{m_k}\right) / \tau\right)}{\sum_{n \in \mathcal{N}, n \neq m} \exp \left(\cos \left(z_{m_i}, z_{n_k}\right) / \tau\right)}, \quad
    \mathcal{L}=\lambda \ell_{i, k}+(1-\lambda) \ell_{k, i}.
\end{equation}
This stage unifies the modalities, embedding image-domain clinical features directly into the aligned manifold.

\textbf{Stage III: End-to-End Analysis from Undersampled k-Space.} In the final stage, the pretrained k-space encoder $\mathcal{E}_k$ is fine-tuned to perform downstream clinical tasks using \textbf{only undersampled k-space measurements} $X^u_k$. We attach lightweight, task-specific decoders to predict phenotype regression, disease classification, and anatomical segmentation from the information-rich latent space. To empirically validate the geometric completeness of the restored latent space, we also evaluate k-MTR's reconstruction ability. By attaching an adaptive image-domain decoder directly to the k-space encoder (similar to AUTOMAP~\cite{zhu2018automap}), k-MTR maps undersampled frequencies to fully sampled images without an explicit inverse Fourier transform.

\section{Dataset and Implementation}

\textbf{Datasets.} The open source literature lacks a foundation-scale type acquired k-space data with labels and annotations. To validate k-space learning, we simulate 42,000 $2D+t$ cardiac MRI scans (6 SAX, 3 LAX; $128\times128\times50$) from the UK Biobank~\cite{petersen2015uk}. We introduce a synthetic phase via Gaussian-smoothed $B_0$ field variation~\cite{brown1999mri} before applying a discrete Fourier transform. We apply a Cartesian undersampling mask~\cite{ahmad2015vista} across all tasks. Specifically, we use acceleration factors of $R = 2, 4, 8,$ and $16$ for phenotype prediction; $R = 4$ for both classification and reconstruction; and a more aggressive $R = 8$ for segmentation, in order to thoroughly evaluate the capability of k-MTR under highly undersampled conditions. Evaluation uses a strictly held-out 1,000-subject test set. For downstream targets, we utilize 12 continuous phenotypes derived from quality-controlled segmentation maps~\cite{bai2018automated}, alongside three disease classifications (coronary artery disease (CAD), high blood pressure, hypertension) following~\cite{zhang2025vita}.

\textbf{Implementation.} Models are implemented in PyTorch on an NVIDIA A100. Stage I MAEs use 6 encoder and 2 decoder layers (1024-dim), a $(5, 8, 8)$ patch size, 70\% masking, and batch size 2. Stage II employs gradient checkpointing for a contrastive batch size of 256, mapping 1025-dim tokens to 128-dim via two-layer MLPs ($\tau=0.1, \lambda=0.5$). In Stage III, the pretrained encoder is fully fine-tuned with task-specific heads: 256-dim MLPs for regression (batch 16) and classification (batch 32), and a 576-dim UNETR~\cite{zhou2023unetr} for segmentation (batch 2). The reconstruction decoder mirrors Stage I, trained from scratch (batch 2).

\textbf{Baselines.} To isolate the efficacy of k-MTR's latent restoration, we evaluate robust baselines. For regression, we compare against ResNet-50~\cite{he2016resnet}, ViT~\cite{dosovitskiy2020image}, and MAE~\cite{zhang2024whole}. ResNet-50 is tested on fully-sampled images (upper bound), artifact-corrupted undersampled images (R=4) (ResNet-50$^u$), and undersampled zero-filled k-space (ResNet-50$_k^u$). Comparing the unaligned counterpart, MAE$_k^u$ (which omits Stages I-II), against k-MTR highlights the necessity of our cross-modal alignment. For classification, we use image-trained ViT and MAE. Segmentation baselines include fully-sampled nnU-Net~\cite{isensee2021nnunet}, undersampled nnU-Net$^u$, and LI-Net~\cite{schlemper2018linet} which is designed for corrupted undersampled images. 

\section{Results}

\begin{figure}[!t]
\center
\includegraphics[width=\linewidth]{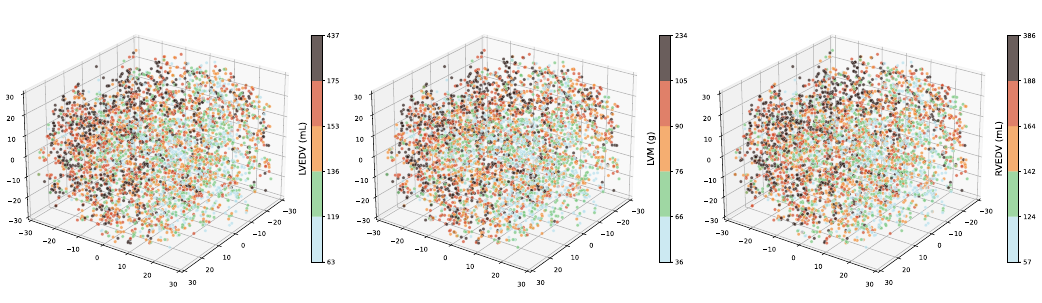}
\caption{3D t-SNE visualization of the representations after alignment, colored by ground-truth phenotype groups.}
\label{fig:tsne}
\end{figure}
\textit{\textbf{Semantic Feature Clustering in Latent Space.}} We first evaluate k-MTR's representational quality to empirically validate the restored latent space. As shown in Fig.~\ref{fig:tsne}, we visualize the latent embeddings of 10000 subjects using 3D t-SNE to evaluate semantic coherence. When color-coded by key cardiac phenotypes (LVEDV, LVM, and RVEDV), the embeddings form meaningful clusters. This separation confirms that k-MTR successfully captures complex spatial and temporal phenotypic variability directly from the undersampled k-space.

\textit{\textbf{Competitive Clinical Downstream Fidelity Direct from Undersampled k-Space.} }We then evaluate k-MTR's ability to extract physiologically meaningful labels natively from undersampled frequencies.

\begin{table}[!t]
    \centering
    \footnotesize
    \setlength{\tabcolsep}{3pt} 
    \caption{Mean absolute error $\downarrow$ for phenotype prediction. Image-based baselines are in \colorbox{gray!30}{gray}. RN50: ResNet-50. Best are in \colorbox{seagreen!50}{dark green}, second and third in \colorbox{seagreen!25}{lighter green}.}
    \label{tab:regression}
    \begin{tabular*}{\textwidth}{@{\extracolsep{\fill}} lccccccc}
    \toprule
    \multirow{2}{*}{Phenotype}
    & \multicolumn{3}{c}{Fully-sampled} 
    & \multicolumn{4}{c}{Undersampled R=4} \\
    \cmidrule(lr){2-4} 
    \cmidrule(lr){5-8} 
    & \cellcolor{gray!30}{RN50}
    & \cellcolor{gray!30}{ViT}
    & \cellcolor{gray!30}{MAE}
    & \cellcolor{gray!30}{RN50$^u$}
    & RN50$_k^u$
    & MAE$_k^u$
    & k-MTR \\
    
    \midrule
    LVEDV (mL)    & \cellcolor{seagreen!50} $6.58$ & $11.53$ & $9.03$ & \cellcolor{seagreen!25} $7.35$ & $9.57$ & $12.88$ & \cellcolor{seagreen!25} $8.15$ \\
    LVSV (mL)     & \cellcolor{seagreen!50} $5.68$ & $6.93$ & \cellcolor{seagreen!25} $5.86$ & $6.55$ & $7.03$ & $8.42$ & \cellcolor{seagreen!25} $6.50$ \\
    LVEF (\%)     & \cellcolor{seagreen!50} $2.95$ & $3.91$ & \cellcolor{seagreen!25} $3.20$ & $3.30$ & $3.58$ & $6.09$ & \cellcolor{seagreen!25} $3.14$ \\
    LVM (g)       & \cellcolor{seagreen!50} $5.51$ & $7.85$ & \cellcolor{seagreen!25} $5.87$ & \cellcolor{seagreen!25} $7.01$ & $8.00$ & $10.09$ & $7.20$ \\
    
    \midrule
    RVEDV (mL)    & \cellcolor{seagreen!50} $9.24$ & $13.75$ & \cellcolor{seagreen!25} $9.49$ & \cellcolor{seagreen!25} $10.49$ & $11.33$ & $14.76$ & $10.60$ \\
    RVESV (mL)    & \cellcolor{seagreen!25} $6.15$ & $8.40$ & \cellcolor{seagreen!50} $6.07$ & $7.35$ & $7.63$ & $8.71$ & \cellcolor{seagreen!25} $6.98$ \\
    RVSV (mL)     & \cellcolor{seagreen!25} $7.58$ & $9.51$ & \cellcolor{seagreen!50} $7.42$ & $8.06$ & $8.41$ & $9.64$ & \cellcolor{seagreen!25} $7.86$ \\
    RVEF (\%)     & \cellcolor{seagreen!25} $3.30$ & $4.17$ & \cellcolor{seagreen!25} $3.62$ & $3.56$ & $3.66$ & $5.97$ & \cellcolor{seagreen!50} $3.28$ \\
    
    \midrule
    LASV (mL)     & \cellcolor{seagreen!25} $4.66$ & $5.55$ & \cellcolor{seagreen!50} $4.10$ & $5.41$ & $5.84$ & $5.78$ & \cellcolor{seagreen!25} $5.34$ \\
    LAEF (\%)     & \cellcolor{seagreen!50} $4.28$ & $5.24$ & \cellcolor{seagreen!25} $4.29$ & $4.82$ & $5.74$ & $6.71$ & \cellcolor{seagreen!25} $4.72$ \\
    
    \midrule
    RASV (mL)     & \cellcolor{seagreen!25} $5.80$ & $7.44$ & \cellcolor{seagreen!50} $5.47$ & $7.08$ & $8.01$ & $7.40$ & \cellcolor{seagreen!25} $7.07$ \\
    RAEF (\%)     & \cellcolor{seagreen!25} $5.09$ & $6.09$ & \cellcolor{seagreen!50} $5.05$ & \cellcolor{seagreen!25} $5.55$ & $6.53$ & $6.56$ & $5.74$ \\
    
    \bottomrule
    \end{tabular*}
\end{table}

\begin{table}[!t]
    \centering
    \footnotesize 
    \setlength{\tabcolsep}{3pt} 
    \caption{Disease classification performance. Fully-sampled image baselines are in \colorbox{gray!30}{gray}; k-MTR with undersampled k-space. Positive class ratios are shown as percentages of the cohort. AP: average precision. Best results are \textbf{bold}, second \underline{underlined}.}
    \label{tab:classification}
    \begin{tabular*}{\textwidth}{@{\extracolsep{\fill}} ccccccc}
    \toprule
    Disease & Method & AUC-ROC$\uparrow$ & F1 Score$\uparrow$  & Recall$\uparrow$ & Precision$\uparrow$ & AP$\uparrow$ \\
    \midrule
     
    \multirow{3}{*}{\begin{tabular}[c]{@{}c@{}}CAD \\ (7.4\%)\end{tabular}} 
    & \cellcolor{gray!30}{ViT} & $0.682$ & $0.176$ & $\textbf{0.842}$ & $0.098$ & $0.126$ \\
    & \cellcolor{gray!30}{MAE} & $\underline{0.708}$ & $\underline{0.215}$ & $0.200$ & $\textbf{0.233}$ & $\underline{0.154}$ \\
    & k-MTR (R=4) & $\textbf{0.737}$ & $\textbf{0.282}$ & $\underline{0.500}$ & $\underline{0.197}$ & $\textbf{0.234}$ \\
    
    \midrule
    \multirow{3}{*}{\begin{tabular}[c]{@{}c@{}}High Blood \\ Pressure \\ (25.8\%)\end{tabular}} 
    & \cellcolor{gray!30}{ViT} & $0.677$ & $0.461$ & $0.652$ & $\underline{0.357}$ & $0.355$ \\
    & \cellcolor{gray!30}{MAE} & $\underline{0.691}$ & $\underline{0.469}$ & $\underline{0.680}$ & $\textbf{0.358}$ & $\underline{0.369}$ \\
    & k-MTR (R=4)& $\textbf{0.697}$ & $\textbf{0.474}$ & $\textbf{0.781}$ & $0.340$ & $\textbf{0.386}$ \\
    
    \midrule
    \multirow{3}{*}{\begin{tabular}[c]{@{}c@{}}Hypertension \\(20.8\%)\end{tabular}} 
    & \cellcolor{gray!30}{ViT} & $0.698$ & $0.406$ & $\underline{0.638}$ & $0.298$ & $\underline{0.360}$ \\
    & \cellcolor{gray!30}{MAE} & $\textbf{0.717}$ & $\textbf{0.432}$ & $\textbf{0.724}$ & $\underline{0.308}$ & $\textbf{0.378}$ \\
    & k-MTR (R=4)& $\underline{0.710}$ & $\underline{0.417}$ & $0.603$ & $\textbf{0.319}$ & $0.356$ \\
    
    \bottomrule
    \end{tabular*}
\end{table}

\textbf{Phenotype Prediction:} As shown in Tab.~\ref{tab:regression}, k-MTR closely approaches the performance of image-based upper bound operating on fully-sampled data. The ability to achieve comparable accuracy to the image-domain baseline on key metrics (e.g., LVEDV and LVEF) provides strong empirical evidence that aligned latent space of k-MTR retains essential physiological semantics. Crucially, k-MTR yields substantial improvements over the unaligned MAE$_k^u$ baseline (which omits Stage II alignment). This confirms that cross-modal contrastive learning successfully embeds localized anatomical awareness into the k-space encoder.

\textbf{Disease classification:} As shown in Tab.~\ref{tab:classification}), k-MTR performs on par with fully-sampled image-based ViT and MAE, notably achieving an AUC of $0.737$ for CAD. Reaching diagnostic equivalence without requiring explicit image reconstruction highlights the semantic richness of the restored latent space.


\begin{figure}[!t]
\center
\includegraphics[width=\linewidth]{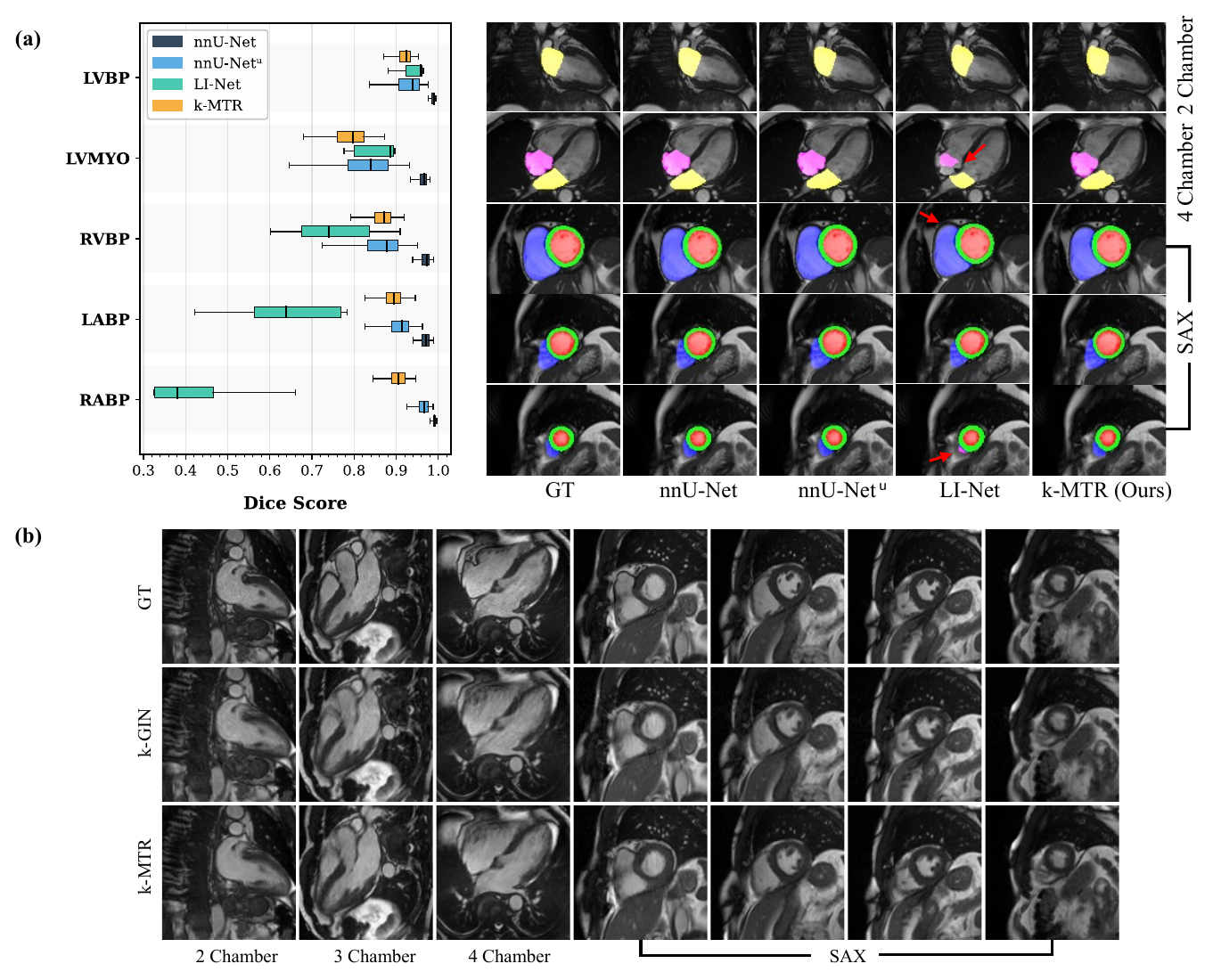}
\caption{\textbf{Segmentation and Reconstruction Results.} \textbf{(a)} Dice Score and example segmentation maps overlayed on fully-sampled images. k-MTR (R=8) is compared against the fully-sampled upper bound (nnU-Net) and undersampled (R=8) image-based baselines (nnU-Net$^u$, LI-Net). \textbf{(b)} Image reconstructions from undersampled k-space compared to k-GIN.}
\label{fig:seg_recon}
\end{figure}

\textbf{Segmentation:} k-MTR exhibits high-precision performance when operating directly on undersampled k-space tokens, achieving an average foreground Dice score of 0.85 at an acceleration factor of R = 8. In contrast, LI-Net struggles to extract reliable semantic information from corrupted images (Fig.~\ref{fig:seg_recon}(a)).

\textbf{Validation Check via Reconstruction:} To assess geometric integrity, we map the undersampled k-space embeddings back to the spatial domain. k-MTR shows on-par results (38.18 $dB$ PSNR) compared to a reconstruction-specific method k-GIN~\cite{pan2023global} model (38.30 $dB$), as shown in Fig.~\ref{fig:seg_recon} (b). By mapping undersampled k-space to image domain directly without IFT, k-MTR learns to implicitly restores omitted anatomical geometries within its manifold.

\begin{figure}[!t]
\center
\includegraphics[width=0.9\linewidth]{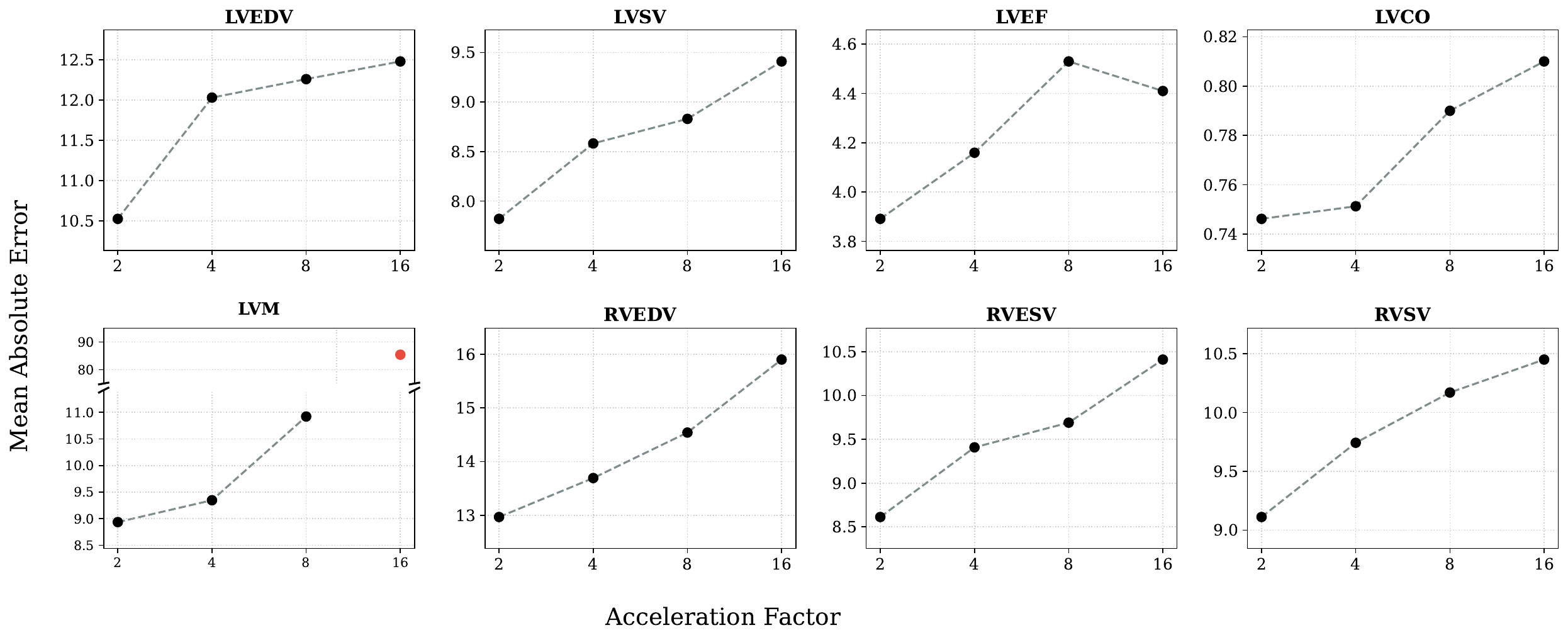}
\caption{Ablation of k-MTR's phenotype prediction accuracy across varying undersampling factors, with 16x prediction failures highlighted in red.}
\label{fig:ablation}
\end{figure}

\textbf{Performance Robustness Across Acceleration Factors.} To assess the extensibility of k-MTR, we evaluate phenotype regression performance under varying k-space undersampling ratios (2x, 4x, 8x, and 16x). As shown in Fig.~\ref{fig:ablation}, performance exhibits a degradation as information loss increases. While k-MTR maintains stable and competitive accuracy under moderate regimes (2x to 8x), the extreme 16x setting causes prediction failures for specific phenotypes (red points). This confirms that the 4-fold setting adopted in the main experiments is a representative and practically relevant choice rather than a limiting case.

\section{Conclusion}

In this work, we introduce k-MTR, a framework that successfully extracts physiologically meaningful representations from undersampled k-space. By establishing a shared frequency-spatial latent space, k-MTR explicitly restores omitted anatomical geometries directly within its manifold, bypassing explicitly solving the inverse problem for cardiac downstream analysis. This cross-modal alignment enables highly competitive, multi-task performance across phenotype regression, disease classification, and anatomical segmentation directly from undersampled k-space.

Future work will extend k-MTR to prospectively acquired multi-coil datasets and systematically evaluate its robustness across diverse sampling patterns and acceleration factors. Existing open-source k-space datasets (e.g., OCMR~\cite{chen2020ocmr}, CMRxRecon~\cite{wang2024cmrxrecon}) lack the clinical annotations necessary for downstream tasks beyond image reconstruction. We plan to annotate these datasets and acquire additional in-house data to validate and expand k-MTR’s potential in a multi-coil setting. We hope this work encourages the community to release diverse, well-annotated multi-coil datasets and provides a robust architectural blueprint for task-aware cardiac MRI workflows that operate directly on undersampled frequency measurements.


\begin{credits}
\subsubsection{\ackname} This research has been conducted using the UK Biobank Resource under Application Number 87802. This work is funded by the European Research Council (ERC) project Deep4MI (884622). Dr. Sevgi Gokce Kafali has been sponsored by the Alexander von Humboldt Foundation.



\end{credits}

%
%
%
\newpage
\bibliographystyle{splncs04_unsrt}
\bibliography{ref_miccai2026}
\end{document}